# Predicting Lung Nodule Malignancies by Combining Deep Convolutional Neural Network and Handcrafted Features

Shulong Li[1], Panpan Xu[2], Bin Li[1], Liyuan Chen[3], Zhiguo Zhou[3], Hongxia Hao[4], Yingying Duan[1], Michael Folkert[3], Jianhua Ma[1], Shiying Huang[5], Steve Jiang[3], and Jing Wang[3*]

[1] School of Biomedical Engineering, Guangdong Provincial Key Laboratory of Medical Image Processing, Southern Medical University, Guangzhou, 510515, China
[2] Longgang District People's Hospital, Shenzhen, 518172, China
[3] Department of Radiation Oncology, University of Texas Southwestern Medical Center, Dallas, 75235, USA
[4] School of Computer Science and Technology, Xidian University, Xi'an, 710071, China.
[5] School of Traditional Chinese Medicine, Southern Medical University, Guangzhou, 510515, China

## Abstract

To predict lung nodule malignancy with a high sensitivity and specificity, we propose a fusion algorithm that combines handcrafted features (HF) into the features learned at the output layer of a 3D deep convolutional neural network (CNN). First, we extracted twenty-nine handcrafted features, including nine intensity features, eight geometric features, and twelve texture features based on grey-level co-occurrence matrix (GLCM) averaged from five grey levels, four distances and thirteen directions. We then trained 3D CNNs modified from three 2D CNN architectures (AlexNet, VGG-16 Net and Multi-crop Net) to extract the CNN features learned at the output layer. For each 3D CNN, the CNN features combined with the 29 handcrafted features were used as the input for the support vector machine (SVM) coupled with the sequential forward feature selection (SFS) method to select the optimal feature subset and construct the classifiers. The fusion algorithm takes full advantage of the handcrafted features and the highest level CNN features learned at the output layer. It can overcome the disadvantage of the handcrafted features that may not fully reflect the unique characteristics of a particular lesion by combining the intrinsic CNN features. Meanwhile, it also alleviates the requirement of a large scale annotated dataset for the CNNs based on the complementary of handcrafted features. The patient cohort includes 431 malignant nodules and 795 benign nodules extracted from the LIDC/IDRI database. For each investigated CNN architecture, the proposed fusion algorithm achieved the highest AUC, accuracy, sensitivity, and specificity scores among all competitive classification models.

**Keywords:** Lung Nodule Malignancy, convolutional neural network, handcrafted feature, fusion algorithm, Radiomics

*Corresponding author.

E-mail address: jing.wang@utsouthwestern.edu (Jing Wang)

# Ⅰ. INTRODUCTION

Lung cancer is the leading cause of cancer-related death in the United States and China. Early detection and diagnosis improve the prognosis for patients with early stage lung cancer treated with surgical resection. The landmark national lung screening trial (NLST) has shown that low-dose computed tomography (LDCT) screening reduces lung cancer mortality by 20% compared to chest radiography (National Lung Screening Trial Research *et al.*, 2011). As more evidence on the benefits of LDCT screening emerges, the U.S. Preventive Services Task Force has recommended "annual screening for lung cancer with LDCT in adults aged 55 to 80 years who have a 30 pack-year smoking history and currently smoke or have quit within the past 15 years" . The Centers for Medicare & Medicaid Services (CMS) have also determined that the evidence is sufficient to add annual screening for lung cancer with LDCT for appropriate beneficiaries. While LDCT screening has demonstrated a 20% survival benefit over chest radiography, the overall false positive rate in NLST was high (26.6%), and the positive predictive value was low (3.8%) (National Lung Screening Trial Research *et al.*, 2011; Pinsky *et al.*, 2015). False positive tests may lead to anxiety, unnecessary and potentially harmful additional follow-up diagnostic procedures, and associated healthcare costs. A reliable strategy is needed to reduce false-positive rates, unnecessary biopsies, and ultimately, patient morbidity and healthcare costs.

To reduce the high false-positive rate in LDCT lung cancer screening, the American College of Radiology developed a new classification scheme named Lung CT Screening Reporting and Data System (Lung-RADS) that 1) increases the size threshold to classify a nodule as positive from 4 mm to 6 mm and 2) requires growth for pre-existing nodules. While applying the Lung-RADS criteria to the NLST data greatly reduced the false-positive rate, it also reduced the baseline sensitivity by nearly 9%, adversely affecting the benefit of LDCT screening to reduce mortality (Pinsky *et al.*, 2015). Radiomics-based approaches present a promising way for lesion malignancy classification (Zhang *et al.*, 2017; Sutton *et al.*, 2016; Gillies *et al.*, 2016; Hao *et al.*, 2018). By extracting and analyzing large amounts of quantitative features from medical images, radiomics can build a predictive model by machine learning algorithms to support clinical decisions. However, even the model that achieved the highest accuracy (80.12%) using a 10-folder cross validation still had a relatively low sensitivity at 0.58. These findings indicate the difficulty of predefining quantitative features that fully reflect the unique characteristics of a particular lesion (Audrey G. Chung, 2015). Thus, developing an effective model based on other input is needed so that only patients with a high probability of developing malignancies undergo additional imaging and invasive testing.

Deep learning has achieved great success in various applications in computer vision (Ding and Tao, 2018) and medical imaging processing and analysis (Fu *et al.*, 2018; Huang *et al.*, 2018b; Zhu *et al.*, 2018a; Parisot *et al.*, 2018). In a recent lung cancer detection challenge organized by Kaggle, most top-scored models were based on a deep convolutional neural network (CNN). Unlike handcrafted feature-based classifiers (Zhou *et al.*, 2017; Liu *et al.*, 2017; Wimmer *et al.*, 2016; Lian *et al.*, 2016; Vallières *et al.*, 2015; Parmar *et al.*, 2015; Namburete *et al.*, 2015), CNN-based classifiers (Kooi *et al.*, 2017; Shin *et al.*, 2016; Roth *et al.*, 2016; Zhu *et al.*, 2018b) use the original images as input and learn features automatically to classify, eliminating the need to extract predefined features. In general, to obtain a good classification performance, CNN requires a large scale annotated dataset to learn the representative nature of a lesion by training a large number of parameters. Many successful applications of CNN have used more than 100,000 samples, such as ImageNet with millions of images (Krizhevsky *et al.*, 2017), the skin cancer dataset with 129,450 images (Esteva *et al.*, 2017), and the retinal fundus photographs dataset with 128,175 images (Gulshan *et al.*, 2016). For many other medical problems, obtaining such a large annotated dataset is still challenging. This challenge is commonly surmounted through transfer learning, which fine-tunes a CNN model pre-trained on a large scale dataset (Kermany *et al.*, 2018; Shin *et al.*, 2016; Akcay *et al.*, 2018). However, for medical imaging with the three dimensional (3D) tensor form, such as CT, transfer learning is not optimal, as most of large datasets are 2D, and there lacks a large scale 3D dataset with pre-trained 3D CNN architecture. Using a conventional classifier (such as SVM) with the CNN features as input is another common technique to improve the performance of CNNs (Shen *et al.*, 2015; Shen *et al.*, 2017). This technique is available for medical imaging with the 3D tensor form, however it can't solve the challenge intrinsically because the natural representation of a lesion can't be learned well by CNNs without a large scale annotated training dataset.

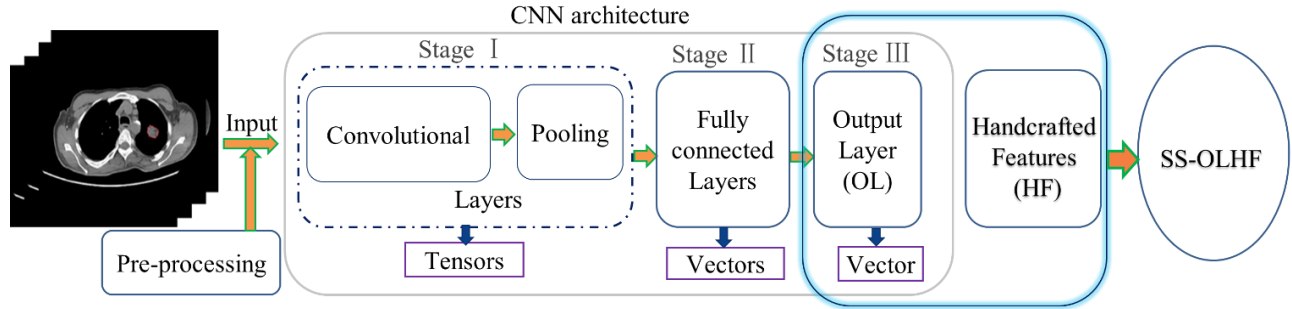

Fig. 1. Illustration of the proposed SS-OLHF algorithm by fusing the highest level CNN representation learned at the output layer (OL) of a 3D CNN into the handcrafted features (HF). In the CNN architecture, Stage I mainly includes convolutional layers and pooling layers with tensor output; Stage II mainly includes hidden fully connected layers with vector output; Stage III is the output layer with vector output.

Currently, most models use either handcrafted features or features learned based on CNN alone. Combining the knowledge extracted by these two methods could enhance a predictive model's performance. On one hand, a combining algorithm could overcome the disadvantage of the handcrafted features' inability to fully reflect the unique characteristics of a particular lesion by combining the intrinsic CNN features. On the other hand, it could alleviate the requirement of a large scale annotated dataset for the CNN because of the complement of handcrafted features for the CNN features. In general, the architecture of a typical CNN (Fig. 1) is structured as a series of stages: 1) convolutional layers and pooling layers with tensor output; 2) hidden fully connected layers with vector output; and 3) an output layer with vector output (LeCun *et al.*, 2015). In most existing combination methods, the representation learned at the final hidden fully connected layer is combined into the handcrafted features, improving the performance of both the handcrafted features and the CNN (Antropova *et al.*, 2017). However, as pointed out by Y. LeCun and G. Hinton, "deep-learning methods are representation-learning methods with multiple levels of representation, obtained by composing simple but non-linear modules that each transform the representation at one level (starting with the raw input) into a representation at a higher, slightly more abstract level" (LeCun *et al.*, 2015). This indicates that the representation learned at the output layer is at a higher level and more abstract than the representation learned at the final hidden fully connected layer. Thus, a fusion algorithm could achieve better performance by combining handcrafted features into the CNN representation learned at the output layer, instead of the final hidden fully connected layer. To the best of our knowledge, this is the first attempt to explore a fusion algorithm between the CNN features at the output layer and the handcrafted features.

Specifically, we propose a fusion algorithm (SS-OLHF) (Fig. 1) that combines the highest level CNN representation learned at the output layer (OL) of a 3D CNN into the domain knowledge, i.e. handcrafted features (HF), using the support vector machine (SVM) coupled with the sequential forward feature selection method (SFS) to select the optimal feature subset and construct the final classifier. The proposed fusion algorithm could lead to better performance in differentiating malignant and benign lung nodules for LDCT lung cancer screening.

## Ⅱ. Materials

We downloaded the Lung Image Database Consortium and Image Database Resource Initiative (LIDC/IDRI) (Armato *et al.*, 2011) (http://www.via.cornell.edu/lidc) to evaluate the proposed fusion classifiers. This dataset includes 1,010 cases, each of which includes images from a clinical thoracic CT scan and an associated XML file that records the annotations from four radiologists. 7,371 lesions were marked "nodule" by at least one of the four radiologists, and 2,669 of those nodules had sizes equal to or larger than 3 mm and were rated with malignancy suspiciousness from 1 to 5 (1 indicates the lowest malignancy suspiciousness, and 5 indicates the highest malignancy suspiciousness).

We considered all nodules with sizes equal to or larger than 3 mm. In total, 2,340 nodules were considered. For each nodule, the malignancy suspiciousness rate was the average value of all rates given by all radiologists who outlined the nodule. By removing ambiguous nodules with malignancy suspiciousness rated at 3, we obtained a total of 431 malignant nodules (average rating>3) and 795 benign nodules (average rating <3) to evaluate our models' performance. Finally, for each nodule, the 3D nodule region of interest (ROI) was extracted based on the contour given by the radiologist who gave the malignancy suspiciousness rate closest to the average.

## Ⅲ. Methods

### 3.1. The workflow of the fusion algorithm SS-OLHF

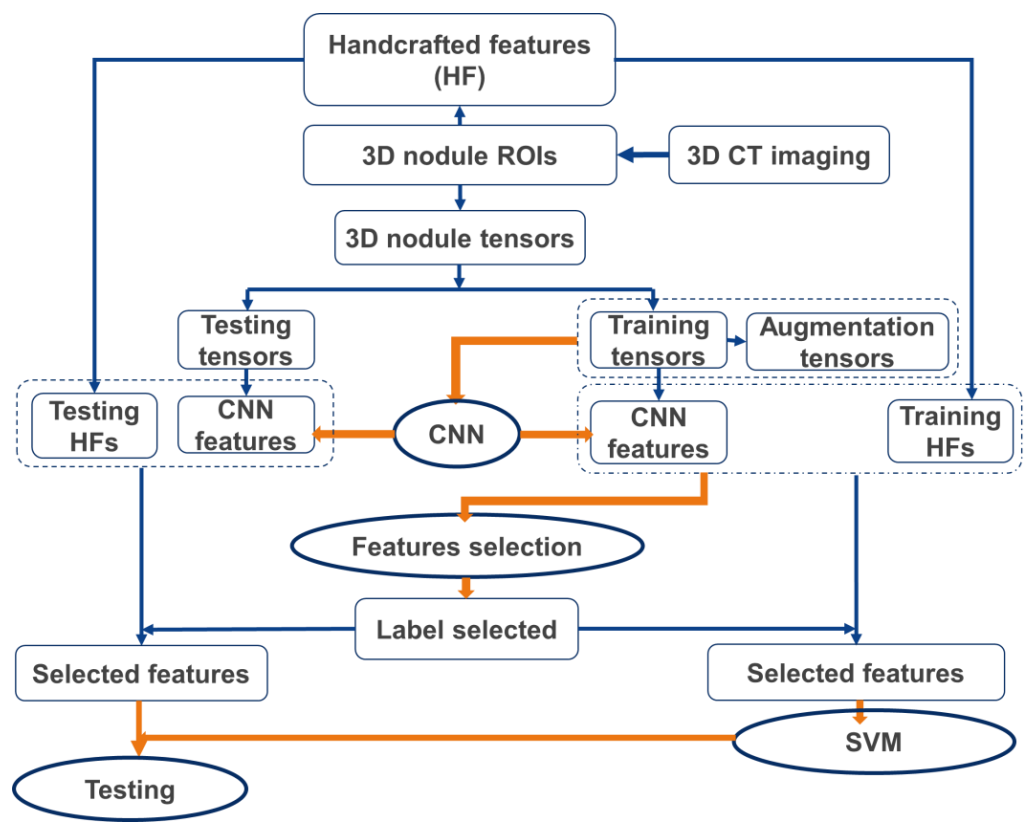

Fig. 2. The overall workflow of the fusion algorithm SS-OLHF.

The overall workflow of the fusion algorithm SS-OLHF is illustrated in Fig. 2. For each nodule, the handcrafted features were extracted and the nodule tensor was constructed based on the segmented 3D nodule ROI. The augmentation processing was performed on the training nodule tensors by rotating and flipping. The 3D tensor, including training tensors and augmentation tensors with the same size, were used as the input for training the 3D CNN. Then, the CNN features learned at the output layers were extracted based on the trained CNN. We obtained the fusion features by combining the CNN features into the handcrafted features. We used SVM coupled with SFS to select the optimal fusion feature subset based on the training samples. Finally, SVM with radial basis function kernel was trained on the training samples using the optimal fusion feature subset and then used as the classifier on the testing samples.

*3.2. Handcrafted feature extraction*

In this study, we combined the handcrafted features into the highest level CNN features learned at the output layer with high abstract. Imaging features, including intensity, geometric and texture features, extracted from contoured nodules (Zhou *et al.*, 2017) were used to construct the fusion algorithm. The following intensity features were extracted based on the intensity histogram: minimum, maximum, mean, standard deviation, sum, median, skewness, kurtosis, and variance. Geometric features associated with a nodule were volume, major diameter, minor diameter, eccentricity, elongation, orientation, bounding box volume, and perimeter. To obtain high level texture features, 360 gray-level co-occurrence matrices (GLCMs) (Davis L.S., 1979) were first constructed based on five grey levels 8, 16, 32, 64, 128, four distances 1, 2, 3, 4 and thirteen directions in 3D linear space, denoted as [0 1 0], [-1 1 0], [-1 0 0], [-1 -1 0], [0 1 -1], [0 0 -1], [0 -1 -1], [-1 0 -1], [1 0 -1], [-1 1 -1], [1 -1 -1], [-1 -1 -1], [1 1 -1]. The following texture features were obtained from the GLCM by averaging the above 360 GLCMs: energy, entropy, correlation, contrast, texture variance, sum-mean, inertia, cluster shade, cluster prominence, homogeneity, max-probability, and inverse variance.

*3.3. Nodule tensor construction for CNNs*

All 3D nodule ROIs were interpolated to a fixed resolution with 0.5 mm/voxel along three axes using a 3D spline interpolation method. Training data were augmented by 2D rotation of [90°, 180°, 270°] along all three axes and flipping along all three coordinate planes. Subsequently, for each interpolated nodule ROI and augmentation data, we constructed the 3D nodule tensor whose center the tumors were located in and whose periphery zeros were filled in. Each nodule tensor had the same size of 105×97×129, determined by the biggest nodule size among all interpolated nodule ROIs and augmentation data. We used these nodule tensors from all interpolated nodule ROIs and augmentation data as the input for all CNNs to train the CNN architectures and extract the CNN features.

*3.4. Convolutional neural network*

*3.4.1. 3D CNN architectures*

Because nodule imaging has an intrinsic 3D tensor structure, the 3D CNN architectures with the 3D nodule tensors as input were trained to extract the CNN features for the proposed fusion algorithm. This study used three 3D CNN architectures, which were modified from two 2D CNN architectures (AlexNet (Krizhevsky *et al.*, 2012) and VGG-16 Net (Simonyan and Zisserman, 2014)) and one recently developed 2D CNN architecture dedicated to classifying lung malignancy (Multi-crop Net (Shen *et al.*, 2017)). Unlike the original 2D CNN architectures, which use 2D convolutional kernels and 2D pooling, our CNN architectures with input in the form of 3D nodule tensors use 3D convolutional kernels to perform 3D convolution in all convolutional layers and 3D max-pooling in all pooling layers. Meanwhile, the 3D CNNs preserve structures, such as layer number at every stage and unit number at every layer (except the output layer with two units), the stride size in most convolutional and pooling layers, and padding processing. The three 3D CNN architectures with corresponding details are shown in Fig. 3 and explained below.

*a) 3D AlexNet* The first 3D CNN architecture is modified from the AlexNet, which achieved significant improvement over other non-deep learning methods for ImageNet Large Scale Visual Recognition (ILSVRC) 2012 (Krizhevsky *et al.*, 2012). This 3D AlexNet architecture includes 5 convolutional layers, three max-pooling layers, two hidden fully connected layers, and one output layer with approximately 113 million free parameters, whose details are shown in Fig. 3(a). The features learned at all convolutional layers and max-pooling layers are in the form of 4D tensors with the size [7 7 9 256] at the final convolutional layer and [4 4 5 256] at the final max-pooling layer. All features learned at the hidden fully connected layers and the output layer are in the form of vectors with the size [1 4096] at the final hidden fully connected layer and [1 2] at the output layer.

*b) 3D VGG-16 Net* The second 3D CNN architecture is modified from the 16-layer 2D CNN developed by the visual geometry group (VGG-16) at the University of Oxford for ImageNet Large Scale Visual Recognition (ILSVRC) 2014 (Simonyan and Zisserman, 2014). This architecture consists of five stacks of convolution-pooling operation, in which each max-pooling layer is tailed by a few convolutional layers (Fig. 3(b)). Overall, it includes thirteen convolutional layers, five max-pooling layers, two hidden fully connected layers, and one output layer with approximately 65 million free parameters. Unlike the original 2D CNN architecture that used all 2D kernels with size 3 and stride 1 in convolutional layers, the first convolutional layer uses 3D kernels with size 11 and stride 4, so that the large input size (105×97×129) works with the available memory of our computer with the modified 3D VGG-16. The features learned at the final convolutional layer, final max-pooling layer, final hidden fully connected layer, and output layer are in the forms of a 4D tensor with size [2 2 3 512], a 2D tensor with size [2 512], a vector with size [1 4096], and a vector with size [1 2], respectively.

*c) 3D Multi-crop Net* The third 3D CNN architecture (Fig. 3(c)) is based on a recently developed multi-crop CNN dedicated to lung malignancy classification (Shen *et al.*, 2017). The multi-crop CNN extracts salient nodule information using a multi-crop pooling strategy that crops center regions from convolutional feature maps and then applies max-pooling at different times. The multi-crop CNN outperforms other state-of-the-art models for classifying lung nodule malignancies, so we chose it as an improved 3D CNN architecture to evaluate our fusion algorithm. In this 3D CNN architecture, which has approximately 0.5 million free parameters, the multi-crop pooling strategy is used after the first convolutional layer, and there are two additional convolutional layers, two max-pooling layers, one hidden fully connected layer, and one output layer. Like the 3D CNN modified from VGG-16 Net, the kernel size is 11 and the stride is 4 at the first convolutional layer. The features learned at the final convolutional layer, final pooling layer, final hidden fully connected layer, and the output layer are in the form of a 4D tensor with size [4 4 5 64], a 4D tensor with size [2 2 3 64], a vector with size [1 32], and a vector with size [1 2], respectively.

*3.4.2. CNN training procedures*

For binary classification, the output of the CNN is a two dimensional vector $(y_0, y_1)$ for each sample with label $q$, where $q$ equals either 0 or 1. The softmax of $(y_0, y_1)$, defined as

$$p_i = \frac{\exp(y_i)}{\exp(y_0)+\exp(y_1)}, i = 0, 1, \tag{1}$$

indicates the probability distribution over the two classes. The networks were trained by minimizing the loss

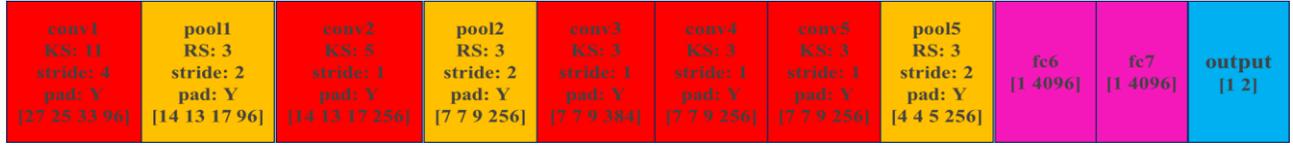

a) 3D Alex Net

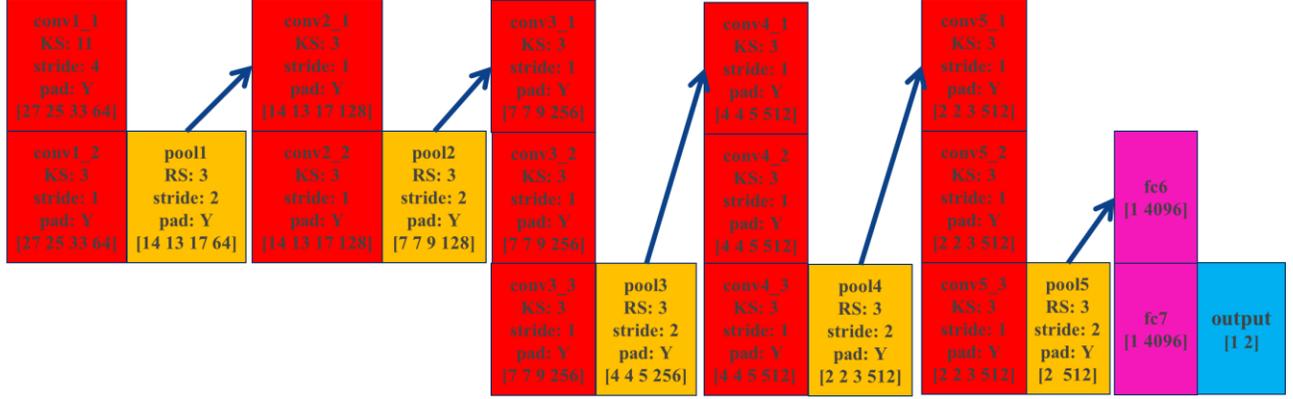

b) 3D VGG-16 Net

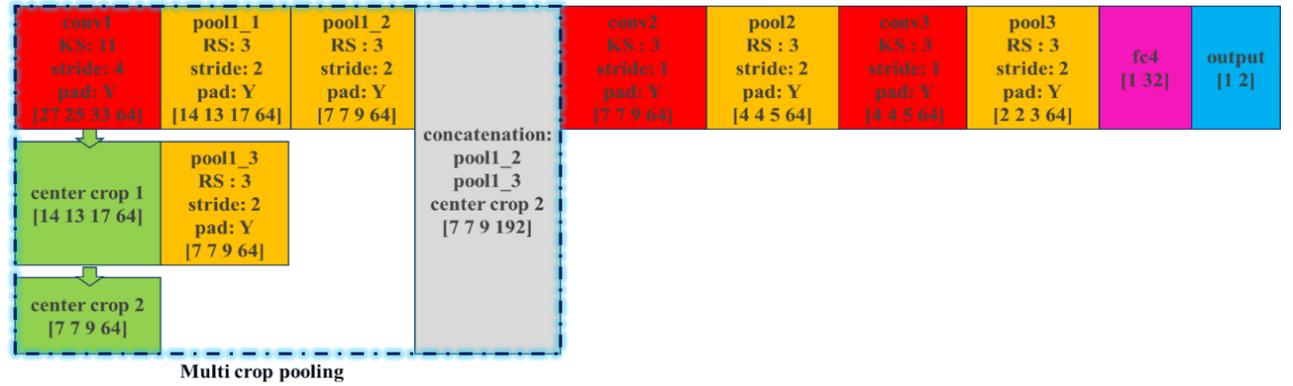

c) 3D Multi-crop Net

Fig. 3. Three 3D CNN architectures modified from two widely used 2D architectures and one 3D CNN dedicated to lung nodule malignancy classification. The vector at bottom of each rectangle indicates the size of the output tensor at this layer for a patient case, with the last number being the unit number. The workflow is from top to bottom and left to right on adjacent rectangles, and along the arrow direction on non-adjacent rectangles. KS: kernel size; RS: region size.

function defined by averaging the cross entropy along each batch with size $N$, as follows:

$$\text{Loss} = \frac{1}{N}\sum -(q\log p_1 + (1-q)\log p_0). \qquad (2)$$

We used the Adam optimization algorithm, based on a first-order gradient (Kingma and Ba, 2014), to optimize the objective function (2). We used minibatch Stochastic Gradient Descent (SGD) to compute the gradient in small batches for the available memory of our computer. The batch size was set to 70. Then, we used the standard backpropagation to adjust weights in all layers. We used Layer (Channels) normalization, introduced by Ba et al. (Jimmy Lei Ba, 2016), to normalize the input x to one nonlinear output based on the mean $\mu$ and standard deviation $\sigma$ over all channels (units) in a layer, as follows,

$$y_{LN}(\mathrm{x}; W, \gamma, \beta) = g\left(\frac{xW - \mu(xW)}{\sigma(xW)}\gamma + \beta\right), \tag{3}$$

where g was the standard rectified linear unit (ReLU), $W$ was the learned weight in this layer, and the extra parameters $\gamma$ and $\beta$ need to be trained during training the CNN. Dropout regularization was used for all hidden fully connected layers with the dropout ratio set to 0.9. The changing learning rate was initially set to 0.005 for the first epoch, 0.001 from the second epoch to the fourth epoch, 0.0005 from the fifth epoch to the eighth epoch, and $10^{-4}$ after the ninth epoch. The training processing stopped automatically when the loss function value achieved 0.01 and iterative steps achieved 100 epochs for each CNN. To deal with the data imbalance problem in training CNNs, we selected part of the majority augmentation data rather than all majority augmentation data to obtain the balanced binary training samples.

*3.5. Feature selection and classification*

The fusion features (Table Ⅰ) comprise 29 handcrafted features, as described in section 3.2, and 2 CNN features learned at the output layer before softmax operating, which are the highest level representation learned by CNN. For each testing sample, one of these 2 CNN features is directly connected with the prediction probability belonging to positive class, which is denoted as CNN featureP, and the other one is directly connected with the prediction probability belonging to negative class, which is denoted as CNN featureN. Because of the redundancy and similarity among these fusion features, feature selection is needed to improve the model performance. As a traditional feature selection method, the sequential forward feature selection (SFS) method (Kohavi and John, 1997) coupled with the SVM classifier (SS) was used to select the feature subset. We used the area under the receive operating characteristic (ROC) curve (AUC) by a 5-fold cross validation method in the training dataset as the criterion to select the optimal feature subset. Finally, the SVM based classifier with RBF kernel was further trained on the training dataset and classified the testing dataset using the selected optimal feature subset as input.

Ⅳ. Experimental Setup

*4.1. Methods for comparative testing*

For each CNN architecture investigated, we first compared the original 2D CNN and the modified 3D CNN with our proposed fusion algorithm SS-OLHF. We also compared our algorithm with other two conventional CNN feature based methods that utilize the features learned at the final hidden fully connected layer (FFL).

*1) S-FFL.* One common CNN feature based method first extracts the CNN features learned at FFL and then uses a third classifier to perform the classification (Shen *et al.*, 2015; Shen *et al.*, 2017). In this method, the 3D CNN features learned at FFL were extracted first, the variance analysis method selected the CNN feature subset to be used as input of a third classifier by removing features with a variance smaller than the mean variance, and SVM with RBF kernel was used to train the classifier and perform the classification.

*2) S-FFLHF.* Conventional fusion algorithms for combining handcrafted features (HF) and CNN extracted features usually use the CNN features learned at the FFL (Antropova *et al.*, 2017), rather than the output layer. We compared our fusion algorithm with this fusion strategy that combines the CNN features learned at FFL into 29 handcrafted features. We used ReliefF (Kononenko *et al.*, 1997), the classical feature selection method, to obtain the optimal feature set based on the training dataset. The SVM model with an RBF kernel was used to construct the classifier.

TABLE Ⅰ. THE FUSION FEATURES USED IN OUR FUSION ALGORITHMS

| Intensity features | Geometry features | Texture features | CNN features |
|---|---|---|---|
| Minimum | Volume | Energy | CNN featureP |
| Maximum | Major diameter | Entropy | CNN featureN |
| Mean | Minor diameter | Correlation | |
| Stand deviation | Eccentricity | Contrast | |
| Sum | Elongation | Texture Variance | |
| Median | Orientation | Sum-Mean | |
| Skewness | Bounding box volume | Inertia | |
| Kurtosis | Perimeter | Cluster Shade | |
| Variance | | Cluster Prominence | |
| | | Homogeneity | |
| | | Max-Probability | |
| | | Inverse Variance | |

Additionally, we compared our fusion algorithm with three other state-of-the-art methods that use different data forms as input. Firstly, the original experimental method of Multi-crop Net that used the nodule patch without segmentation as input (oriMulti-crop) was compared with our proposed fusion algorithm using the segmented 3D nodule ROI as input. To show that the CNN features can be as the complement of handcrafted features to improve the classification accuracy, the same classifier and feature selection processing using handcrafted features (SS-HF) only as input was compared with our fusion algorithms. The third method compared with our fusion algorithms is the support tensor machine with the 3D nodule tensor as input. These three compared approaches are described in detail as following:

*1) oriMulti-crop.* Different from the our proposed fusion algorithm using segmented 3D nodule ROIs, the original experimental method of Multi-crop net (oriMulti-crop) used 3D nodule patches without ROI segmentation as the input (Shen *et al.*, 2017). The experimental results show that the oriMulti-crop outperforms other state-of-art models in classifying lung nodule malignancies. We compared results from the oriMulti-crop with the 3D nodule patches covering the nodule as the input with our fusion algorithm, using the nodule patch size 32*32*32 with the same resolution and the same training protocols for both algorithms.

*2) SS-HF.* We also evaluated our fusion algorithm by comparing it with the classification using just the 29 handcrafted features with the same feature selection processing (SFS) and classifier training method (SVM).

*3) STM.* The support tensor machine (STM) (Tao *et al.*, 2007), which uses a high order tensor as input, is a common tensor space model and has been applied successfully to pedestrian detection, face recognition, remote sensors, and medical imaging analysis (Guo *et al.*, 2016; Biswas and Milanfar, 2017; Li *et al.*, 2018). The 3D STM uses the 3D nodule tensors directly as input, so it doesn't require the extraction of predefined features. Moreover, the parameters needed to train are much smaller than with CNN. When we compared STM with our proposed algorithm, we used all original 3D nodule tensors without augmentation tensors to train the STM model to perform classification.

## 4.2. Experimental setting

To solve the class imbalance problem, for the vector space models, we used the Synthetic Minority Over-sample Technique (SMOTE) (Chawla *et al.*, 2002) to generate a synthetic vector sample based on minority class information to augment the decision region of the minority class using the K-nearest neighborhood (KNN) graph based on Euclidean distance. For all CNNs, we handled the imbalance problem in training processing by randomly selecting part of the augmentation tensors in the majority class and all of the augmentation tensors in the minority class to obtain a balanced training dataset. For the STM, we used the original nodule tensors to train the model, randomly selecting part of the nodule tensors in the majority class and all of the nodule tensors in the minority class to obtain a balanced training dataset.

We employed a 5-fold cross validation method to evaluate the performance of the different classifiers. All samples were randomly partitioned into 5 subsets with a size of either 245 or 246. These five subsets were fixed for each method investigated in this study, where one subset was used as the testing subset and the rest were used as training data. Within the training samples, we employed the 5-fold cross validation method during the training process to select the feature subset which consisted of all features selected for at least four folds. The classification models were then trained on all samples in the training set using this feature subset. Finally, the trained models performed classification on the testing subset.

For each testing subset, we calculated the AUC, classification accuracy, sensitivity, and specificity. We used the average results and standard deviations from one round of 5-fold experiments, as the evaluation criteria. Additionally, the ISO-accuracy lines (Nicolas Lachiche, 2003) are used, defined as

$$\text{TPr} = \frac{P(N)}{P(P)} FPr + a, \qquad (4)$$

where TPr and FPr are the true positive rate and false positive rate, respectively, P(N) and P(P) are the probability of the positive class and the probability of the negative class, respectively, and $a$ is a variable. The ISO-accuracy lines are a family of lines that are parallel, i.e. that have the same slope. The point on the ROC curve where the ISO-accuracy line is tangent obtains the optimal probability threshold to obtain the optimal accuracy. Therefore, we also used the ISO-accuracy line tangent to the ROC curve to evaluate the performance of all classification approaches in this study.

# Ⅴ. Results

## 5.1. Comparison with CNN-based methods

The results comparing our fusion algorithm with different CNN architectures and other methods based on CNN are summarized in Table Ⅱ. For each CNN architecture, the 3D CNN obtained better results than the 2D CNN. SVM with the CNN features learned at the final hidden fully connected layer as input (S-FFL) improved the performance of 3D CNNs in terms of AUC, classification accuracy, sensitivity, and specificity. The conventional fusion algorithm combining the CNN features learned at the final hidden fully connected layer into handcrafted features (S-FFLHF) obtained better results than S-FFL in most of cases, but not for all four evaluated metrics of the three CNN architectures. Our proposed fusion algorithm obtained the best performance for every CNN architecture. The best AUC (=0.9307) was obtained based on VGG-16 Net, the best sensitivity (82.60%) was obtained based on AlexNet and Multi-crop Net, and the best specificity (91.95%) and classification accuracy (88.66%) were obtained based on AlexNet.

TABLE Ⅱ. PERFORMANCE OF DIFFERENT PREDICTIVE MODELS BASED ON CNNs

| Methods | | 2D | 3D | S-FFL | S-FFLHF | SS-OLHF |
|---|---|---|---|---|---|---|
| AlexNet | AUC (%) | 88.87 ±2.87 | 90.56 ±1.81 | 91.13 ±1.70 | 91.25 ±1.87 | 93.03 ±2.92 |
| | ACC (%) | 84.67 ±2.01 | 84.99 ±1.86 | 86.05 ±1.87 | 86.88 ±0.40 | 88.66 ±3.72 |
| | SEN (%) | 78.60 ±7.42 | 80.88 ±5.15 | 81.99 ±4.28 | 81.06 ±1.81 | 82.60 ±8.09 |
| | SPE (%) | 88.10 ±2.84 | 87.30 ±1.99 | 88.29 ±1.22 | 90.11 ±0.52 | 91.95 ±1.58 |
| VGG16 Net | AUC (%) | 86.22 ±3.26 | 90.34 ±4.00 | 91.76 ±3.01 | 91.05 ±3.17 | 93.07 ±2.33 |
| | ACC (%) | 85.73 ±2.45 | 86.13 ±2.84 | 87.03 ±2.47 | 85.94 ±0.45 | 87.60 ±2.91 |
| | SEN (%) | 80.05 ±8.08 | 80.29 ±4.36 | 80.71 ±7.20 | 80.84 ±2.95 | 82.85 ±7.97 |
| | SPE (%) | 88.81 ±3.88 | 89.30 ±2.20 | 90.60 ±2.08 | 88.81 ±1.21 | 90.14 ±3.71 |
| Multi-crop Net | AUC (%) | 89.18 ±3.14 | 90.48 ±3.51 | 90.86 ±2.95 | 92.70 ±2.34 | 93.06 ±1.92 |
| | ACC (%) | 85.73 ±2.64 | 86.46 ±2.53 | 86.62 ±1.61 | 86.05 ±0.80 | 88.58 ±2.70 |
| | SEN (%) | 80.44 ±6.57 | 81.91 ±6.28 | 82.28 ±5.14 | 80.37 ±2.08 | 82.60 ±6.13 |
| | SPE (%) | 88.70 ±1.79 | 88.93 ±1.88 | 89.06 ±0.73 | 90.37 ±2.71 | 91.82 ±1.86 |

The ROC curves of our proposed algorithm always lie above the ROC curves of the 2D CNN, 3D CNN, and S-FFL, which indicates that our proposed fusion algorithm outperforms the 2D CNN, 3D CNN, and S-FFL for all three CNN architectures (Fig. 4 (a)-(c)). The ISO-accuracy line tangent to ROC for our fusion algorithm is above and to the left of the conventional fusion algorithm for all three investigated CNN architectures (Fig. 4 (a)-(c)). This indicates that our proposed fusion algorithm obtains a better TPr and smaller FPr than the conventional fusion algorithm S-FFLHF, though the ROC curve of our fusion algorithms doesn't always lie above the ROC curve of S-FFLHF.

Our proposed fusion algorithms based on three CNN architectures yield results with small differences (Table II). The fusion algorithm based on Multi-crop net has steadier results than the other algorithms, because its standard deviations (AUC: 1.92; ACC: 2.70; SEN: 6.13; SPE: 1.86) for five independent experiments are the smallest. It obtained results of 0.9306, 88.58%, 82.60%, and 91.82% for AUC, accuracy, sensitivity, and specificity, respectively.

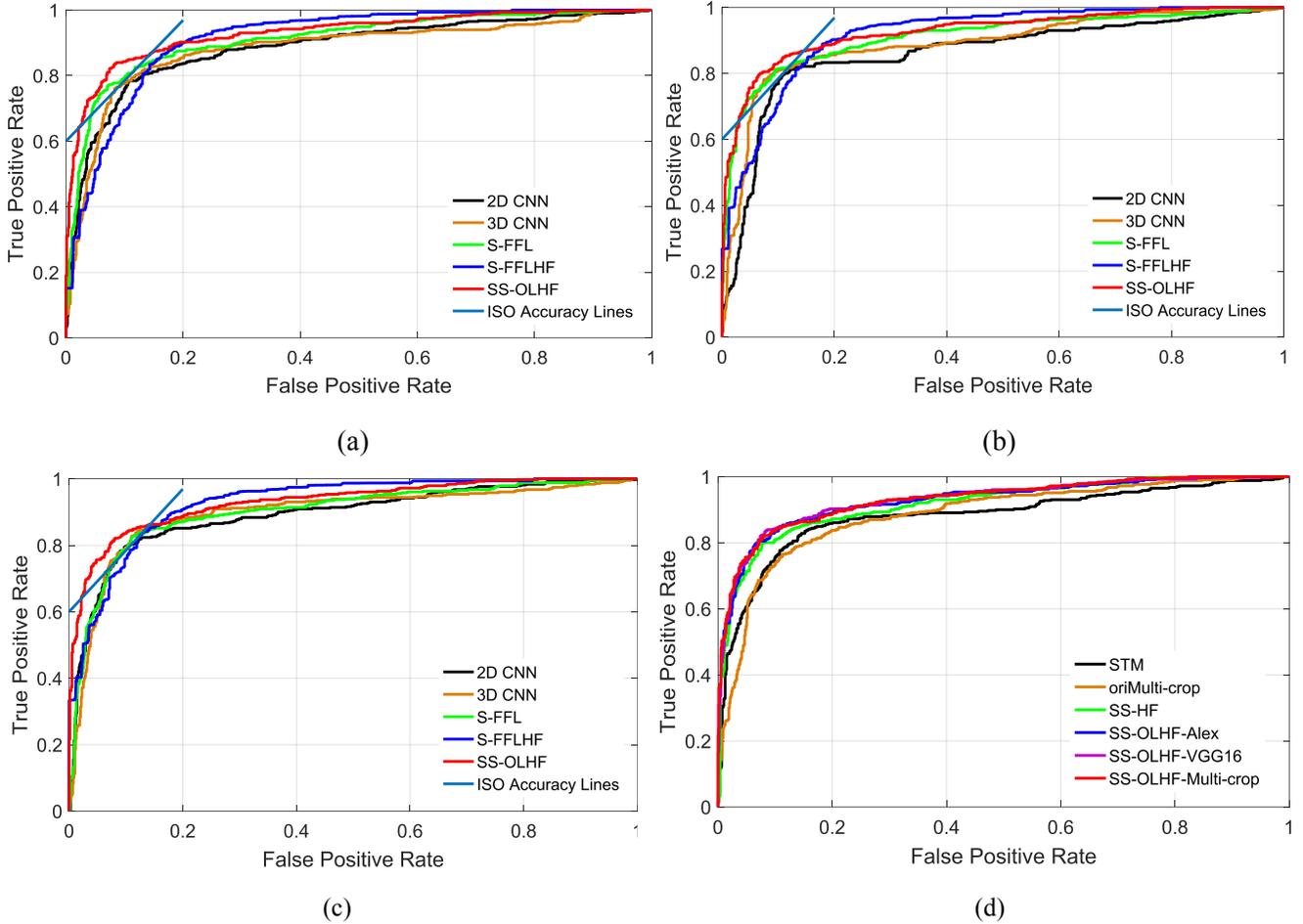

Fig. 4. The ROC curves for the compared methods: (a) five methods based on AlexNet architecture; (b) five methods based on VGG-16 Net architecture; (c) five methods based on Multi-crop Net architecture; (d) six methods, including our three fusion algorithms, based on different CNN architectures and three conventional approaches with different data forms as input.

TABLE Ⅲ. PERFORMANCE OF THREE STATE-OF-THE-ART APPROACHES

| Methods | AUC(%) | ACC(%) | SEN(%) | SPE(%) |
| --- | --- | --- | --- | --- |
| oriMulti-crop | 89.24 ±1.81 | 82.54 ±2.76 | 80.94 ±3.65 | 83.36 ±4.00 |
| SS-HF | 90.45 ±2.58 | 85.62 ±2.37 | 81.21 ±6.20 | 89.56 ±1.17 |
| STM | 88.47 ±4.09 | 84.26 ±3.33 | 83.29 ±8.02 | 84.78 ±2.60 |

TABLE Ⅳ. P-VALUES IN THE UNPAIRED T-TEST BETWEEN THE PERFORMACE OF OUR FUSION ALGORITHM BASED ON MULTI-CROP NET AND THE OTHER THREE METHODS

| Methods | AUC | ACC |
| --- | --- | --- |
| oriMulti-crop | 0.0135 | 0.0161 |
| SS-HF | 0.0493 | 0.0004 |
| STM | 0.0462 | 0.0275 |

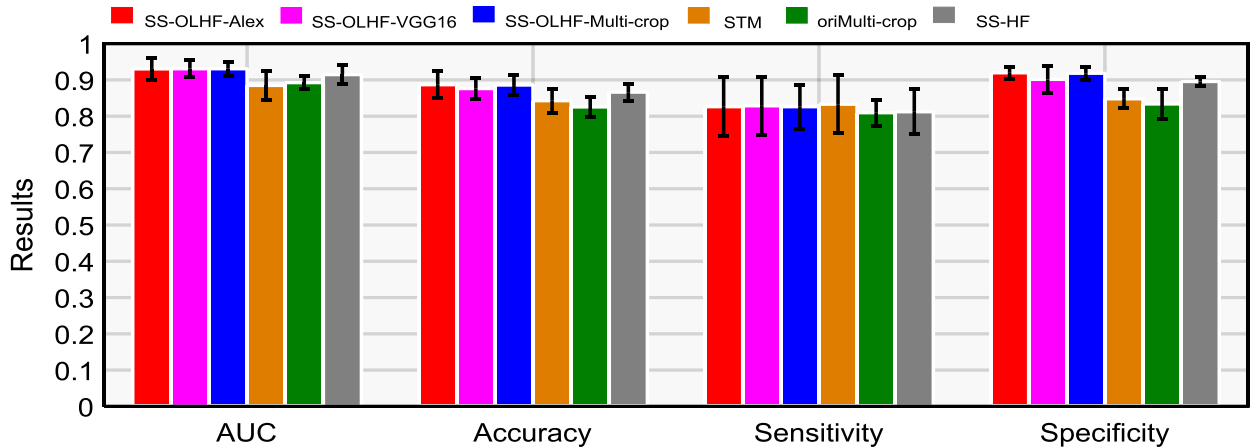

Fig. 5. The comparison results among our three fusion algorithms and other three state-of-the-art approaches.

*5.2. Comparison with other state-of-the-art approaches*

Table Ⅲ shows the results obtained from the three state-of-the-art approaches with different forms of data as input for modeling and testing, as described in section 4.1. All of the fusion algorithms with three CNN architectures outperformed the other three competitive approaches (Fig. 4 (d) and Fig. 5). The corresponding p-values in T-test are reported in Table Ⅳ from the 5 independent experimental results compared with the fusion algorithm based on Multi-crop Net as it is steadiest. These results show that the fusion algorithm based on Multi-crop Net is significantly better than the other three methods, as shown by the p-values < 0.05.

*5.3. Features selected by the fusion algorithms*

The features selected in each testing fold by our fusion algorithm with three CNN architectures are shown in Table Ⅴ. The fusion algorithm based on AlexNet selected 5 optimal features: 2 intensity features, one texture feature, one geometry feature, and one CNN feature. The fusion algorithm based on VGG16 selected 8 features: 4 intensity features, one texture feature, 2 geometry features, and one CNN feature. The fusion algorithm based on Multi-crop Net selected 9 features: 2 intensity features, 4 texture features, 2 geometry features, and one CNN feature. Variance as the intensity feature, contrast as the texture feature, minor diameter as geometry feature, and CNN featureP were selected by all three fusion algorithms, which indicates the complementarity among these three types of handcrafted features and CNN features.

TABLE Ⅴ THE FEATURES THAT ARE SELECTED BY OUR FUSION ALGORITHMS

| The fusion algorithms | Intensity features | Geometry features | Texture features | CNN features |
|---|---|---|---|---|
| SS-OLHF-Alex | Stand deviation<br>Variance | Minor diameter | Contrast | CNN featuresP |
| SS-OLHF-VGG16 | Maximum<br>Stand deviation<br>Sum<br>Variance | Major diameter<br>Minor diameter | Contrast | CNN featuresP |
| SS-OLHF-Multi-crop | Sum<br>Variance | Volume<br>Minor diameter | Contrast<br>Texture Variance<br>Cluster Shade<br>Inverse Variance | CNN featuresP |

# Ⅵ. Discussions and conclusions

Lung cancer screening based on LDCT has shown to reduce the mortality of lung cancer patients in the NLST but false positive rate is very high. CNN has been a powerful tool in many fields. However, it generally requires a large scale annotated dataset to learn the natural representation of an object. The conventional radiomics method, based on handcrafted features, has performed well for many tasks with relatively small sample size. However, the handcrafted features may not fully reflect the unique characteristics of particular lesions. To solve these problems and to develop a highly sensitive and specific model that differentiates between malignant and benign lung nodules, we proposed a fusion algorithm that incorporates the CNN representation learned at the output layer of a 3D CNN into the handcrafted features, using the feature selection method SFS coupled with SVM selects the optimal feature subset and constructs the classifier. Unlike conventional fusion methods that combine the CNN representation learned at the final hidden fully connected layer into the handcrafted features, our proposed fusion algorithm combines the highest level CNN representation learned at the output layer into the handcrafted features.

We investigated three 3D CNN architectures – AlexNet, VGG-16 Net, and Multi-crop Net – modified from the 2D CNN architectures with different parameter values in our proposed fusion algorithm. The experimental results (Table Ⅱ-Ⅳ) show that the fusion algorithm performed best in all competitive approaches. The fusion algorithm based on Multi-crop Net obtained steadier results than our fusion algorithms based on the other two CNN architectures. As the steadiest fusion algorithm (achieving lowest standard deviation of different evaluation criteria), the fusion algorithm based on Multi-crop Net obtained 82.60% and 91.82% for sensitivity and specificity, respectively. The false positive rate is 8.28% and false negative rate is 17.4%, which are relatively low. While both sensitivity and specificity are improved in the proposed fusion algorithm as compared to other state-of-art classifiers, the sensitivity isn't that high (<85%), which might be induced by the imbalanced patients between the classes. One feasible solution to obtain more balanced solution is to develop a multi-objective model where both sensitivity and specificity are considered as the objectives during model optimization (Zhou *et al.*, 2018). The three CNN investigated in this study were proposed in 2012, 2014, and 2017, respectively. The performance of the fusion algorithm could be further improved by using more advanced CNN architectures such as ResNet (He *et al.*, 2015) and DenseNet (Huang *et al.*, 2018a). Furthermore, a prospective study is desired to evaluate the proposed fusion algorithm.

In this study, the following features: variance as the intensity feature, contrast as the texture feature, minor diameter as the geometry feature, and CNN featureP were selected by all our three fusion algorithms based on three CNN architectures, although different optimal feature subsets were obtained by different CNN architectures (Table Ⅴ). These selections are desirable as the first two features selected by all three fusion algorithms indicate the heterogeneity of lesions, the third one indicates the size of lesions, and the CNN featureP learned at the output layer of CNN indicates the intrinsic characteristics. On the other hand, we used the SFS feature selection method coupled with SVM classifier to select the optimal feature subset. The different optimal features were obtained by the fusion algorithms based on the different CNN architectures, which could be mainly induced by the properties of the current feature selection method. Feature selection depends on many factors, such as classifiers, feature set, and selection processing, so that different factors would induce the different selection results. Therefore, more robust feature selection methods are needed to further improve the robustness and performance of the proposed algorithm.

Our proposed fusion algorithms were performed on the nodule ROI with manual segmentation by four radiologists. The manual segmentation is time consuming as well as bringing difference between different radiologists. Many automatic segmentation methods were proposed (Badrinarayanan *et al.*, 2017; Chen *et al.*, 2018). Recently, many researchers focus on the CNN research and disease quantification without segmentation (Tong *et al.*, 2019). These techniques can be incorporated into our fusion algorithm removing the step of manual segmentation.

## Acknowledgement

This work was partly supported by the American Cancer Society (ACS-IRG-02-196), the US National Institutes of Health (5P30CA142543), and the National Natural Science Foundation of China (NSFC, 11771456 and U1708261). The authors would like to thank Dr. Jonathan Feinberg for editing the manuscript.

## References

USPSTF Final Recommendation Statement Lung Cancer: Screening.

Akcay S, Kundegorski M E, Willcocks C G and Breckon T P 2018 Using Deep Convolutional Neural Network Architectures for Object Classification and Detection Within X-Ray Baggage Security Imagery *Ieee T Inf Foren Sec* **13** 2203-15

Antropova N, Huynh B Q and Giger M L 2017 A deep feature fusion methodology for breast cancer diagnosis demonstrated on three imaging modality datasets *Medical physics* **44** 5162-71

Armato S G, McLennan G, Bidaut L, McNitt-Gray M F, Meyer C R, Reeves A P, Zhao B, Aberle D R, Henschke C I and Hoffman E A 2011 The lung image database consortium (LIDC) and image database resource initiative (IDRI): a completed reference database of lung nodules on CT scans *Medical physics* **38** 915-31

Audrey G. Chung M J S, Devinder Kumar, Farzad Khalvati, Masoom A. Haider, Alexander Wong 2015 Discovery Radiomics for Multi-Parametric MRI Prostate Cancer Detection

Badrinarayanan V, Kendall A and Cipolla R 2017 SegNet: A Deep Convolutional Encoder-Decoder Architecture for Image Segmentation *Ieee T Pattern Anal* **39** 2481-95

Biswas S K and Milanfar P 2017 Linear Support Tensor Machine With LSK Channels: Pedestrian Detection in Thermal Infrared Images *Ieee T Image Process* **26** 4229-42

Chawla N V, Bowyer K W, Hall L O and Kegelmeyer W P 2002 SMOTE: synthetic minority over-sampling technique *Journal of artificial intelligence research* **16** 321-57

Chen L, Shen C, Li S, Maquilan G, Albuquerque K, Folkert M R and Wang J 2018 Automatic PET cervical tumor segmentation by deep learning with prior information *Medical Imaging* **1057436**

Davis L.S. J S A, Aggarwal J. 1979 Texture analysis using generalized co-occurrence matrices *Ieee T Pattern Anal* 251-9

Ding C X and Tao D C 2018 Trunk-Branch Ensemble Convolutional Neural Networks for Video-Based Face Recognition *Ieee T Pattern Anal* **40** 1002-14

Esteva A, Kuprel B, Novoa R A, Ko J, Swetter S M, Blau H M and Thrun S 2017 Dermatologist-level classification of skin cancer with deep neural networks *Nature* **542** 115-8

Fu H Z, Cheng J, Xu Y W, Wong D W K, Liu J and Cao X C 2018 Joint Optic Disc and Cup Segmentation Based on Multi-Label Deep Network and Polar Transformation *IEEE transactions on medical imaging* **37** 1597-605

Gillies R J, Kinahan P E and Hricak H 2016 Radiomics: Images Are More than Pictures, They Are Data *Radiology* **278** 563-77

Gulshan V, Peng L, Coram M, Stumpe M C, Wu D, Narayanaswamy A, Venugopalan S, Widner K, Madams T, Cuadros J, Kim R, Raman R, Nelson P C, Mega J L and Webster R 2016 Development and Validation of a Deep Learning Algorithm for Detection of Diabetic Retinopathy in Retinal Fundus Photographs *Jama-J Am Med Assoc* **316** 2402-10

Guo X, Huang X, Zhang L, Zhang L, Plaza A and Benediktsson J A 2016 Support Tensor Machines for Classification of Hyperspectral Remote Sensing Imagery *IEEE Transactions on Geoscience and Remote Sensing* **54** 3248-64

Hao H X, Zhou Z G, Li S L, Maquilan G, Folkert M R, Iyengar P, Westover K D, Albuquerque K, Liu F, Choy H, Timmerman R, Yang L and Wang J 2018 Shell feature: a new radiomics descriptor for predicting distant failure after radiotherapy in non-small cell lung cancer and cervix cancer *Phys Med Biol* **63**

He K M, Zhang X Y, Ren S Q and Sun J 2015 Deep Residual Learning for Image Recognition *arXiv:1512.03385*

Huang G, Liu Z, Maaten L v d and Weinberger K Q 2018a Densely Connected Convolutional Networks *arXiv:1608.06993*

Huang H, Hu X T, Zhao Y, Makkie M, Dong Q L, Zhao S J, Guo L and Liu T M 2018b Modeling Task fMRI Data Via Deep Convolutional Autoencoder *IEEE transactions on medical imaging* **37** 1551-61

Jimmy Lei Ba J R K, Geoffrey E. Hinton 2016 Layer Normalization *arXiv:1607.06450v1*

Kermany D S, Goldbaum M, Cai W, Valentim C C S, Liang H, Baxter S L, McKeown A, Yang G, Wu X, Yan F, Dong J, Prasadha M K, Pei J, Ting M Y L, Zhu J, Li C, Hewett S, Dong J, Ziyar I, Shi A, Zhang R, Zheng L, Hou R, Shi W, Fu X, Duan Y, Huu V A N, Wen C, Zhang E D, Zhang C L, Li O, Wang X, Singer M A, Sun X, Xu J, Tafreshi A, Lewis M A, Xia H and Zhang K 2018 Identifying Medical Diagnoses and Treatable Diseases by Image-Based Deep Learning *Cell* **172** 1122-31 e9

Kingma D and Ba J 2014 Adam: A method for stochastic optimization *arXiv preprint arXiv:1412.6980*

Kohavi R and John G H 1997 Wrappers for feature subset selection *Artificial intelligence* **97** 273-324

Kononenko I, Šimec E and Robnik-Šikonja M 1997 Overcoming the myopia of inductive learning algorithms with RELIEFF *Applied Intelligence* **7** 39-55

Kooi T, Litjens G, van Ginneken B, Gubern-Merida A, Sanchez C I, Mann R, den Heeten A and Karssemeijer N 2017 Large scale deep learning for computer aided detection of mammographic lesions *Medical image analysis* **35** 303-12

Krizhevsky A, Sutskever I and Hinton G E *Advances in neural information processing systems,2012), vol. Series)* pp 1097-105

Krizhevsky A, Sutskever I and Hinton G E 2017 ImageNet Classification with Deep Convolutional Neural Networks *Commun Acm* **60** 84-90

LeCun Y, Bengio Y and Hinton G 2015 Deep learning *Nature* **521** 436-44

Li S L, Yang N, Li B, Zhou Z G, Hao H X, Folkert M R, Iyengar P, Westover K, Choy H, Timmerman R, Jiang S and Wang J 2018 A pilot study using kernelled support tensor machine for distant failure prediction in lung SBRT *Medical image analysis* **50** 106-16

Lian C, Ruan S, Denœux T, Jardin F and Vera P 2016 Selecting radiomic features from FDG-PET images for cancer treatment outcome prediction *Medical image analysis* **32** 257-68

Liu M, Zhang J, Yap P-T and Shen D 2017 View-aligned hypergraph learning for Alzheimer's disease diagnosis with incomplete multi-modality data *Medical image analysis* **36** 123-34

Namburete A I, Stebbing R V, Kemp B, Yaqub M, Papageorghiou A T and Noble J A 2015 Learning-based prediction of gestational age from ultrasound images of the fetal brain *Medical image analysis* **21** 72-86

National Lung Screening Trial Research T, Aberle D R, Adams A M, Berg C D, Black W C, Clapp J D, Fagerstrom R M, Gareen I F, Gatsonis C, Marcus P M and Sicks J D 2011 Reduced lung-cancer mortality with low-dose computed tomographic screening *N Engl J Med* **365** 395-409

Nicolas Lachiche P F 2003 Improving accuracy and cost of two-class and multi-class probabilistic classifiers using ROC curves *PRoceedings of the Twentieth International Conference on Machine Learning (ICML-2003), Washington DC.*

Parisot S, Ktena S I, Ferrante E, Lee M, Guerrero R, Glocker B and Rueckert D 2018 Disease prediction using graph convolutional networks: Application to Autism Spectrum Disorder and Alzheimer's disease *Medical image analysis* **48** 117-30

Parmar C, Grossmann P, Bussink J, Lambin P and Aerts H J 2015 Machine Learning methods for Quantitative Radiomic Biomarkers *Scientific reports* **5** 13087

Pinsky P F, Gierada D S, Black W, Munden R, Nath H, Aberle D and Kazerooni E 2015 Performance of Lung-RADS in the National Lung Screening Trial: a retrospective assessment *Ann Intern Med* **162** 485-91

Roth H R, Lu L, Liu J, Yao J, Seff A, Cherry K, Kim L and Summers R M 2016 Improving Computer-Aided Detection Using  Convolutional Neural Networks and Random View Aggregation *IEEE transactions on medical imaging* **35** 1170-81

Shen W, Zhou M, Yang F, Yang C and Tian J 2015 Multi-scale Convolutional Neural Networks for Lung Nodule Classification *Information processing in medical imaging : proceedings of the ... conference* **24** 588-99

Shen W, Zhou M, Yang F, Yu D, Dong D, Yang C, Zang Y and Tian J 2017 Multi-crop Convolutional Neural Networks for lung nodule malignancy suspiciousness classification *Pattern Recognition* **61** 663-73

Shin H-C, Roth H R, Gao M, Lu L, Xu Z, Nogues I, Yao J, Mollura D and Summers R M 2016 Deep convolutional neural networks for computer-aided detection: CNN architectures, dataset characteristics and transfer learning *IEEE transactions on medical imaging* **35** 1285-98

Simonyan K and Zisserman A 2014 Very deep convolutional networks for large-scale image recognition *arXiv preprint arXiv:1409.1556*

Sutton E J, Dashevsky B Z, Oh J H, Veeraraghavan H, Apte A P, Thakur S B, Morris E A and Deasy J O 2016 Breast cancer molecular subtype classifier that incorporates MRI features *J Magn Reson Imaging* **44** 122-9

Tao D, Li X, Wu X, Hu W and Maybank S J 2007 Supervised tensor learning *Knowledge and Information Systems* **13** 1-42

Tong Y, Udupaa J K, Odhner D, Wu C, Schuster S J and Torigian D A 2019 Disease quantification on PET/CT images without explicit object delineation *Medical image analysis* **51** 169-83

Vallières M, Freeman C, Skamene S and El Naqa I 2015 A radiomics model from joint FDG-PET and MRI texture features for the prediction of lung metastases in soft-tissue sarcomas of the extremities *Physics in medicine and biology* **60** 5471-96

Wimmer G, Tamaki T, Tischendorf J J, Häfner M, Yoshida S, Tanaka S and Uhl A 2016 Directional wavelet based features for colonic polyp classification *Medical image analysis* **31** 16-36

Zhang Y, Oikonomou A, Wong A, Haider M A and Khalvati F 2017 Radiomics-based Prognosis Analysis for Non-Small Cell Lung Cancer *Scientific reports* **7** 46349

Zhou Z, Folkert M, Iyengar P, Westover K, Zhang Y, Choy H, Timmerman R, Jiang S and Wang J 2017 Multi-objective radiomics model for predicting distant failure in lung SBRT *Physics in medicine and biology* **62** 4460

Zhou Z G, Li S L, Qin G G, Folkert M, Jiang S and Wang J 2018 Automatic multi-objective based feature selection for classification *arXiv:1807.03236*

Zhu B, Liu J Z, Cauley S F, Rosen B R and Rosen M S 2018a Image reconstruction by domain-transform manifold learning *Nature* **555** 487-+

Zhu W T, Liu C C, Fan W and Xie X H 2018b DeepLung: Deep 3D Dual Path Nets for Automated Pulmonary Nodule Detection and Classification *arXiv: 1801.09555v1*